\documentclass[conference]{IEEEtran}
\IEEEoverridecommandlockouts
\usepackage{cite}
\usepackage{amsmath,amssymb,amsfonts}
\usepackage{algorithmic}
\usepackage{graphicx}
\usepackage{textcomp}
\usepackage{xcolor}
\usepackage{booktabs}   
\usepackage{multirow}   

\def\BibTeX{{\rm B\kern-.05em{\sc i\kern-.025em b}\kern-.08em
    T\kern-.1667em\lower.7ex\hbox{E}\kern-.125emX}}

\usepackage[export]{adjustbox}
\usepackage{setspace}
    
\begin{document}

\newcommand{\IF}[1]{\textcolor{red}{#1}}
\newcommand{\KS}[1]{\textcolor{blue}{#1}}

\title{VLMs for Videogame Data Annotation}

\author{\IEEEauthorblockN{Katrin Schmid}
\IEEEauthorblockA{\textit{NVIDIA}, Australia \\
kschmid@nvidia.com}
\and
\IEEEauthorblockN{Iuri Frosio}
\IEEEauthorblockA{\textit{NVIDIA}, Italy \\
ifrosio@nvidia.com}
}


\maketitle

\begin{abstract}
Vision Language Models (VLMs) and Artificial
Intelligence (AI) agents have revolutionized how engineers approach
complex problems in real-world applications. Their adoption
in video games is on the other hand limited by the extreme
variability of the synthetic scenarios and their poor compliance
with real-world physics. Here we investigate the use of VLMs for
annotating video game frame sequences with reward signals, a task with several potential applications including, among others, conditioned training and offline reinforcement learning.
We show that VLMs often struggle to answer basic questions
on racing video games (although we observed a similar behavior on other game genres) and discuss countermeasures such as VLM output mixing and prompt optimization. We also show how
input sequence length, resolution, and question batching affect
the annotation quality and its token consumption.
\end{abstract}


\section{Introduction}

\begin{figure*}[h]
    \begin{center}
    \includegraphics[width=0.186\linewidth, cfbox=blue 1pt 1pt]{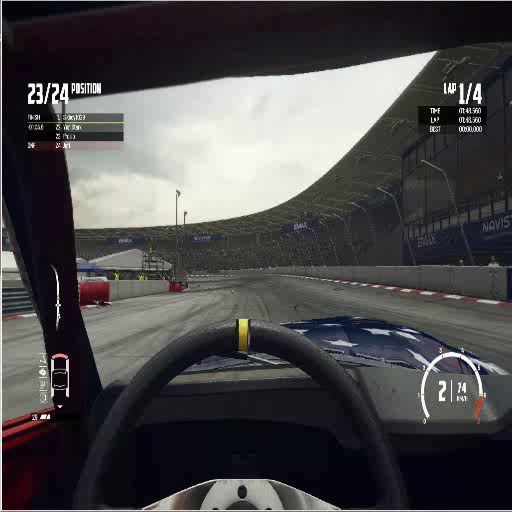}
    \includegraphics[width=0.186\linewidth, cfbox=blue 1pt 1pt]{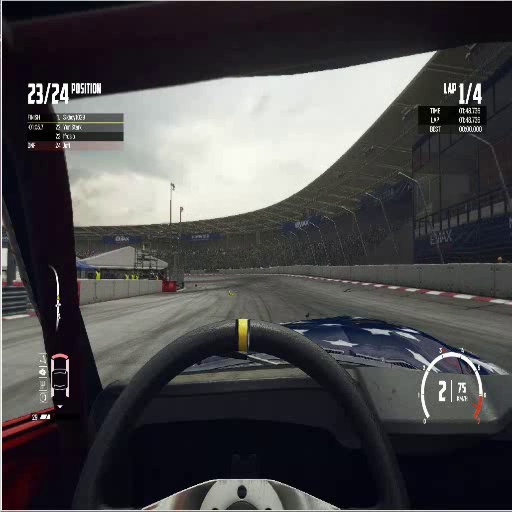}
    \includegraphics[width=0.186\linewidth, cfbox=blue 1pt 1pt]{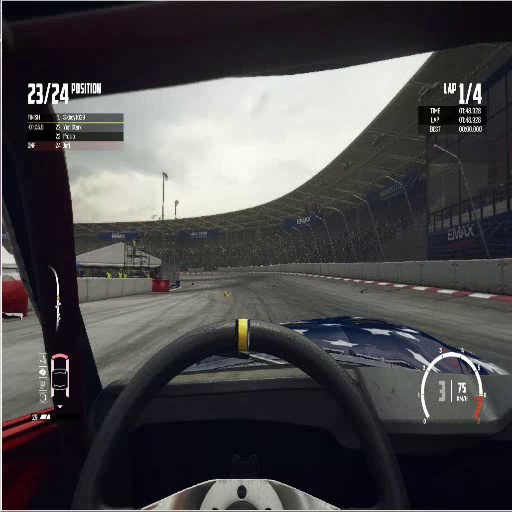}
    \includegraphics[width=0.186\linewidth, cfbox=blue 1pt 1pt]{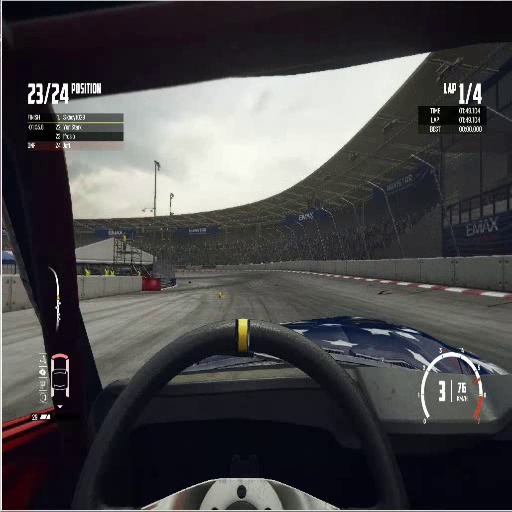}
    \includegraphics[width=0.186\linewidth, cfbox=blue 1pt 1pt]{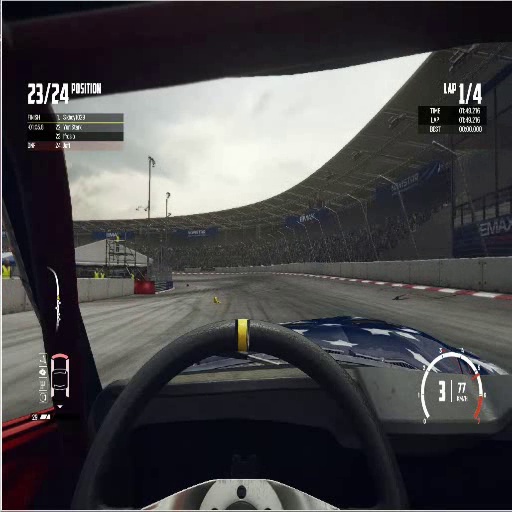}
    \\
    \includegraphics[width=0.186\linewidth, cfbox=cyan 1pt 1pt]{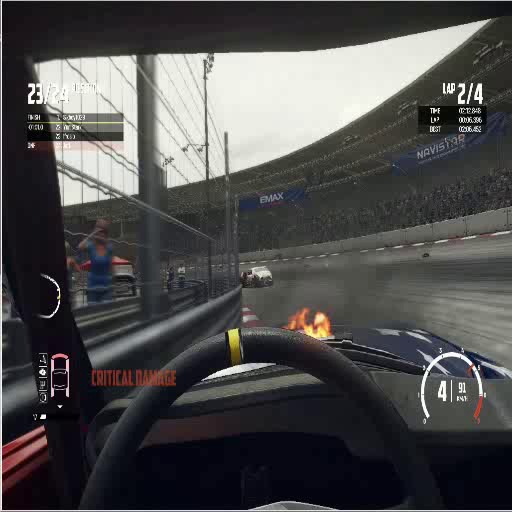}
    \includegraphics[width=0.186\linewidth, cfbox=cyan 1pt 1pt]{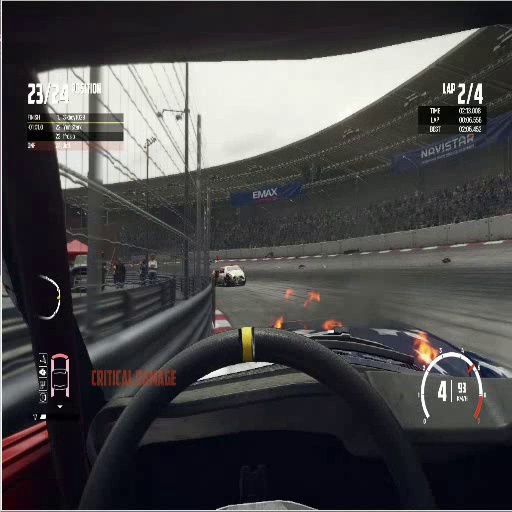}
    \includegraphics[width=0.186\linewidth, cfbox=cyan 1pt 1pt]{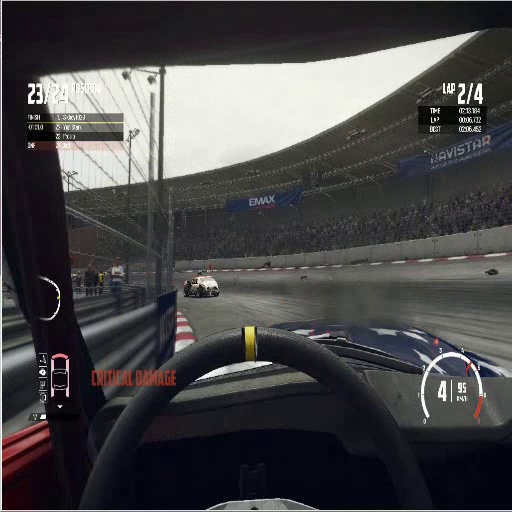}
    \includegraphics[width=0.186\linewidth, cfbox=cyan 1pt 1pt]{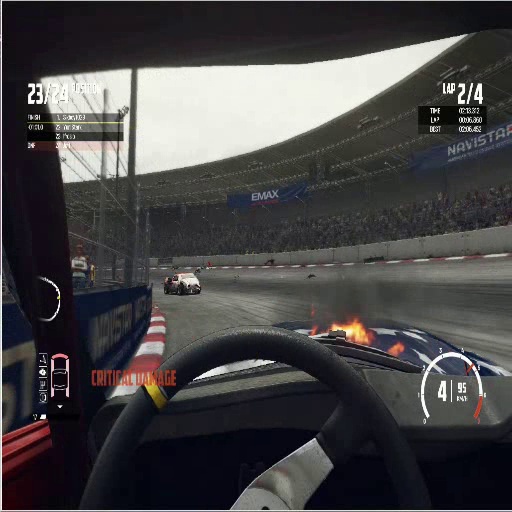}
    \includegraphics[width=0.186\linewidth, cfbox=cyan 1pt 1pt]{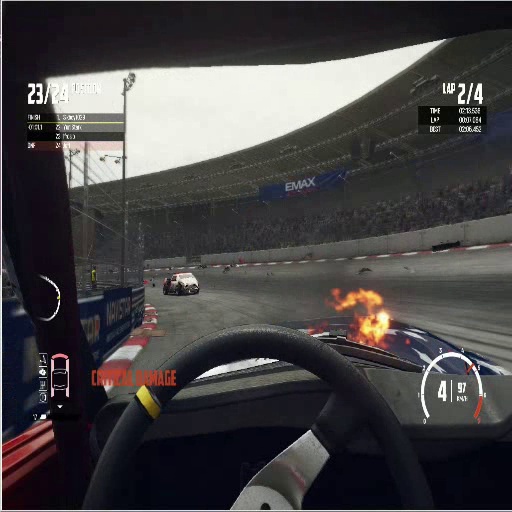}
    \end{center}
    \begin{spacing}{0.8}
    \scriptsize
    \textbf{(1) ChatGPT-5.4, blue sequence. \emph{Is the car in the middle of the road?}} \textcolor{green}{Yes, the car appears to be roughly in the middle of the road, maybe slightly left of center.}\\
    \textbf{(2) Qwen-3.5, blue sequence. \emph{Is the car in the middle of the road?}} \textcolor{red}{Based on the visual evidence in the images, we can determine the car's position on the track. The track has clear boundaries marked by red and white curbing on both the left and right sides. In all five frames, the car is positioned significantly to the left of the track's centerline. It is driving close to the inside edge of the turn, near the red and white curb. This racing line, often called the "apex," is a common technique used by drivers to maintain speed through a corner. It is not the geometric middle of the road. Therefore, the car is not in the middle of the road; it is positioned on the left side of the track as it navigates a turn.}\\
    \textbf{(3) Nemotron2-12B, blue sequence. \emph{Is the car in the middle of the road?}} \textcolor{green}{Yes, the car is in the middle of the road.}\\
    \textbf{(4) ChatGPT-5.4, cyan sequence. \emph{Is the car in the middle of the road?}} \textcolor{red}{Yes. The car ahead is positioned roughly in the middle of the road, slightly left of center.}\\
    \textbf{(5) Qwen-3.5, cyan sequence. \emph{Is the car in the middle of the road?}} \textcolor{green}{Based on the provided images, the car is not in the middle of the road. Here's a breakdown of its position: the viewpoint is from inside the cockpit of the player's car, looking forward down the racetrack. The player's car is positioned very close to the inside wall of the track, which is on the left side of the screen. This is evident from the proximity of the wall and the catch fence. The racing line, or the optimal path through the turn, is generally towards the center and then the outside of the corner. The player's car is on the extreme inside edge. There is another white car visible further ahead on the track, and it is positioned more towards the center of the available track width compared to the player's car. Therefore, the car is on the far left side of the track, hugging the inside wall, not in the middle.}\\
    \textbf{(6) Nemotron2-12B, cyan sequence. \emph{Is the car in the middle of the road?}} \textcolor{red}{Yes, the car is in the middle of the road. The image shows a racing car positioned centrally on the track, with other vehicles visible on either side. This central positioning is typical in racing scenarios, where cars compete for optimal track positioning to gain an advantage over their competitors.}\\
    \textbf{(7) ChatGPT-5.4, cyan sequence. \emph{Does the car hit the lateral barrier?}} \textcolor{green}{Yes. The car is already scraping/pressed against the left-side lateral barrier in these frames.}\\
    \textbf{(8) Qwen-3.5, cyan sequence. \emph{Does the car hit the lateral barrier?}} \textcolor{red}{Based on the provided images, the car does not appear to hit the lateral barrier. Here's a breakdown of the visual evidence: Proximity: The car is driving extremely close to the barrier on the left side of the track. In some frames, it appears to be just inches away. Lack of Contact Indicators: Despite the proximity, there are no visual cues that typically signify an impact in a racing game: There are no sparks flying from the car or the barrier. The car's trajectory does not abruptly change as it would after a collision. It continues to follow the curve of the track smoothly. The "CRITICAL DAMAGE" warning is already present on the screen across all frames, suggesting the car was damaged prior to this specific moment, not from hitting the barrier in these images. The car is navigating a turn at high speed while severely damaged, but it successfully avoids making contact with the wall in this sequence.}\\ 
    \textbf{(9) Nemotron2-12B, cyan sequence. \emph{Does the car hit the lateral barrier?}} \textcolor{red}{Yes, the car hits the lateral barrier. The image shows a car with a "CRITICAL DAMAGE" warning, indicating it has collided with the barrier. The impact has caused visible damage to the vehicle, including a fire on the hood, which is a common result of such collisions in racing games. The car's position and the damage it has sustained clearly demonstrate that it has made contact with the lateral barrier on the track.}
    \end{spacing}
    \caption{Correct (green) and wrong (red) annotation of two Wreckfest sequences with different VLMs.}
    \label{fig:teaser}
\end{figure*}

VLMs and AI agents demonstrate a remarkable understanding of the real world~\cite{reed2022generalist, simateam2025sima2generalistembodied}, enabling applications ranging from robotics~\cite{reed2022generalist} to open-ended embodied  interaction~\cite{wang2024voyager}.
Although agents have been deployed for some video games~\cite{reed2022generalist, wang2024voyager, simateam2025sima2generalistembodied}, AI still struggles with many gaming tasks.
For instance, Fig.~\ref{fig:teaser} shows annotations generated by ChatGPT-5.4~\cite{OpenAI2026ChatGPT54}, Qwen-3.5~\cite{qwen35omni} and Nemotron2-12B on Wreckfest.
The VLMs may: misinterpret the scene visual features (case 2 in Fig.~\ref{fig:teaser}); need a more precise prompt (case 4); hallucinate other vehicles and wrongly assume the car to be in optimal position (case 6); wrongly associate the displayed damage report with a recent impact against the barrier (case 9); or correctly identify the damage as pre-existing yet still failing to detect the barrier contact from visual evidence (case 8).
The reasons for failure are numerous: among them, the visual complexity and large range of variation observed in video games are dominant.
AI models have been mostly trained on real-world data and therefore suffer from a significant domain gap: for instance, video game physics may allow jumping for meters or crashing a car without reporting any damage, which is in contrast with the common-sense inductive bias instilled in AI models.

Despite these difficulties, the potential for AI in video games remains immense: applications range from automatic bots~\cite{10.1111/cgf.15173} to game commentators~\cite{zheng2025surveyAIGGC} and AI coaches. Here we focus on annotation of video game datasets with a set of dense, human-defined rewards that can be used to facilitate offline policy learning~\cite{Sutton1998}, as shown in~\cite{2026vlmconditioned}.
Our contributions are:
\begin{itemize}
\item we compare
several VLMs on Visual Question Answering for Trackmania and Wreckfest game sequences;
\item we introduce a linear mix model to
improve annotation accuracy;
\item we analyze the effects of prompt optimization, input sequence length and resolution, and question batching on accuracy and cost.
\end{itemize}
\section{Related Work}
\label{sec:RelatedWork}

VLMs have been used as data annotators in various contexts. 
In Eureka~\cite{ma2023eureka} a Large Language Model (LLM) generates reward functions for robotics tasks; it outperforms human-engineered rewards without asking for any task-specific prompting or pre-defined reward templates, but requires a programmable environment to score the generated functions.
Here we adopt a simpler procedure where a VLM produces per-frame reward labels directly from rendered frames.
MEGAnno+\cite{Kim2024MEGAnnoAH} proposes a collaborative LLM and human-in-the-loop strategy to help LLMs on complex, sociocultural, or domain-specific contexts; it does not handle visual inputs.
GaLA\cite{Pra2026GaLA} targets video games: it shows that general models struggle with zero-shot interpretation of gaming scenarios and proposes VLM finetuning to close the domain gap of pretrained models on game-specific concepts.
Motif\cite{klissarov2024motif} leverages screen text to annotate data with a LLM and build an intrinsic reward system for reinforcement learning that, paired with the original rewards, leads to a better policy.
Similarly to our Reward Annotation Model (RAM), Snorkel~\cite{ratner2017snorkel} aggregates outputs from raw classifiers through majority vote or a generative label model, but does so without labeled data.
We also notice that agents that directly interact with video games~\cite{simateam2025sima2generalistembodied} have been developed, but their compute cost and latency are strong limitations.

\section{Method}
\label{sec:method}

\begin{table*}
\caption{Premise $P$ and questions $\{Q_j\}_{j=1.. 3}$ for prompting VLMs.}
\label{tab:prompt}
\centering
\scriptsize
\begin{tabular}{p{17.7cm}}
    $P$ You are given a sequence of five frames from the Trackmania video game. They have been captured at 20 frames per second. You are an expert game play analyst and your task is to answer one question about this frame sequence. You must focus on the red-and-white car, the concrete road, the green curbs, and the side walls. Please ignore the grass, the sky, and other stadium elements. Your answer must be a single integer in the 0 to 100 range. Your answer must be 100 if there is clear evidence for a positive answer to the question. If the answer to the question is no, you must output 0. In case of uncertainty, your answer must be a number in the 1 to 99 interval representing your level of confidence. For instance, in case of total uncertainty, you should output 50. You can perform reasoning internally, but you must answer with an integer only. The question is: \\
    \midrule
    $Q_1$ is the red-and-white car in the middle of the road? \\
    $Q_2$ is the red-and-white car moving forward? \\
    $Q_3$ is the red-and-white car hitting the lateral barriers? \\    
\end{tabular}
\end{table*}

\textbf{Dataset annotation.} \label{sec:method_data_plain}
Our scope is to automatically annotate long video game sequences with $N$ expert-defined, dense rewards $\{R_j\}_{j=1..N}$, each associated with a specific question $Q_j$.
We use a premise prompt $P$ to provide context and ask the VLM to output a probability in the $0-100\%$ range to answer $Q_j$.
This is not a limitation for learning as $R_j$ can be arbitrarily scaled in training.
The full prompt for reward $j$ is built concatenating $P$ and $Q_j$; the one adopted in our experiments is shown in 
Table \ref{tab:prompt}.
Reward annotation requires a compromise between the annotation cost (in terms of consumed tokens and their corresponding price), temporal accuracy (some events, like hitting a barrier while driving, take only a few frames, while others, like being in a car accident, can take longer) and the VLM's need for context (e.g., consecutive frames are required to analyze motion). We found annotating sequences of five frames at 20{Hz} to work well for Trackmania.
Therefore, to annotate long clips, we annotate 5-frame sequences with 1 frame overlap using the VLM and interpolate the output probabilities to get a continuous reward signal.

\textbf{Reward Annotation Model.}
Rewards often show some form of correlation: for instance, a car is unlikely to be in the middle of the road ($Q_1$) when it hits a lateral barrier ($Q_3$).
We leverage these correlations to improve the annotation quality through our Reward Annotation Model (RAM).
Unlike Snorkel\cite{ratner2017snorkel}, which aggregates weak labels by voting, RAM learns a simple linear combination of VLM outputs on a small human-annotated reference set and use it to improve the quality of the annotation. We indicate with $A_j(t)$ the ground truth, human-annotated reward for question $j$ at frame $t$ (for the practical details of annotation, see Section \ref{sec:results}).
We indicate with $a^{VLM}_j(t)$ the probability output by a VLM for $Q_j$ on frame $t$.
RAM generates a linear mix of the VLM outputs as:
\begin{equation}
    a^{RAM}_j(t) = \sum\nolimits_{k=1}^N{w_{k,j} a^{VLM}_k(t)} + b_j,
\label{eq:ram}
\end{equation}
where $w_{k,j}$ and $b_j$ are learned independently for each question $Q_j$ by either minimization of an $L2$ loss:
\begin{equation}
    L^j_{L2} = \sum\nolimits_{t=1}^T{(A_j(t) - a^{RAM}_j(t))^2 / T}
    \label{eq:L2}
\end{equation}
or, better, by maximization of a soft $F\beta$ score:

\begin{eqnarray}
TP_j = \sum\nolimits_{t=1}^T{A_j(t) * a^{RAM}_j(t)}/T\\
FP_j = \sum\nolimits_{t=1}^T{(1 - A_j(t)) * a^{RAM}_j(t)}/T\\
FN_j = \sum\nolimits_{t=1}^T{A_j(t) * (1 - a^{RAM}_j(t))}/T\\
F\beta_j = \frac{(1 + \beta^2) \cdot TP_j}{((TP_j + FN_j) \cdot \beta^2 + (TP_j + FP_j) + \epsilon)}\\
L_{F\beta}^j = 1 - F\beta_j,
\end{eqnarray}
where $TP_j$, $FP_j$, and $FN_j$ are the true / false positive / negative rates, $F\beta_j$ is the soft $F\beta$ for question $j$, and $\epsilon=1\cdot10^{-6}$ is a regularization terms to avoid divisions by zero.
Depending on $\beta$, $F\beta$ forces a different precision/recall compromise --- a useful feature for sparse rewards or high cost for reward missing or overestimation.
Here we report results for $F1$ as we found that the $L2$ loss (or $F\beta$ for $\beta \neq 1$) generally achieve worse results.
We optimize with RMSProp, a batch size as large as the entire dataset, and 10000 iterations, returning the RAM model with the best validation.

\textbf{Prompt optimization}
Prompting is a fundamental part of the AI inference chain. Manual prompt optimization was partially effective for our work but time-consuming and hard to handle. We therefore tested the DSPy framework with a Multi-prompt Instruction Proposal Optimizer (MIPROv2)~\cite{opsahl-ong-etal-2024-optimizing}, which compiles LLM pipelines by automatically optimizing prompts from few-shot examples against a target metric, replacing manual prompt engineering.
For optimization we used the same human-annotated dataset used to learn RAM. 
MIPROv2 performs Bayesian optimization over the prompt space; each question $Q_j$ ends up with a premise $P_j$ that differs from the original one and the optimization is model-specific.

\textbf{Question batching}
The compute cost is measured here in terms of tokens.
To minimize the number of calls, one possible strategy is to use a premise $P$ like the one in Table \ref{tab:prompt} while asking the VLM to output a vector that answers all the questions $\{Q_j\}_{j=1...N}$ at the same time.
We refer to this as \emph{batched} configuration.

\section{Results}
\label{sec:results}

\begin{figure}
 \begin{center}
    \includegraphics[width=0.19\linewidth]{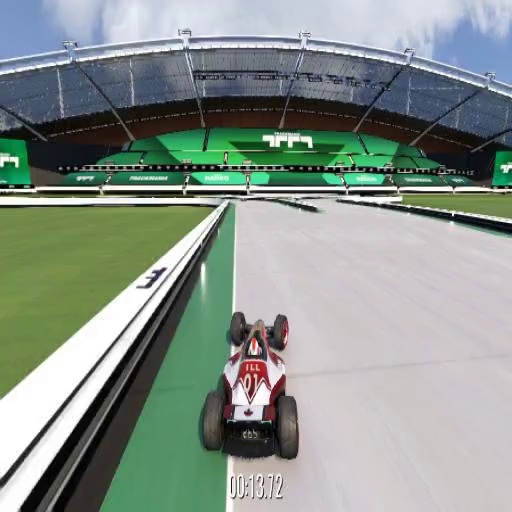}
    \includegraphics[width=0.19\linewidth]{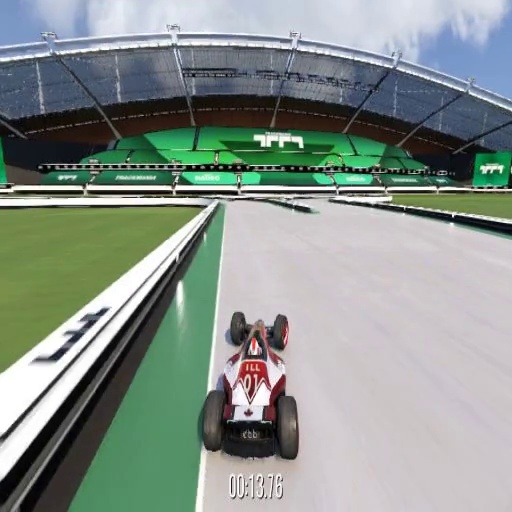}
    \includegraphics[width=0.19\linewidth]{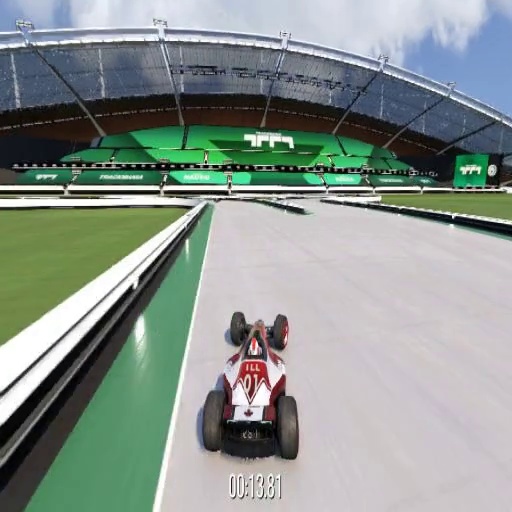}
    \includegraphics[width=0.19\linewidth]{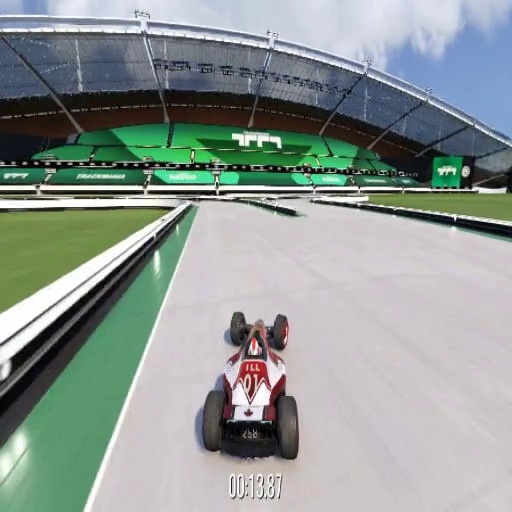}
    \includegraphics[width=0.19\linewidth]{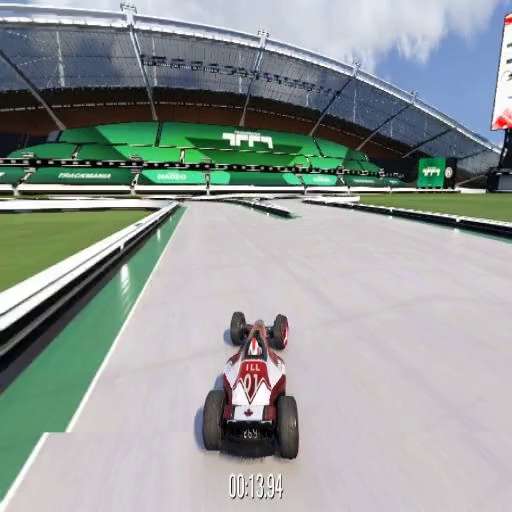}\\
    \vspace{0.1cm}
    \includegraphics[width=0.19\linewidth]{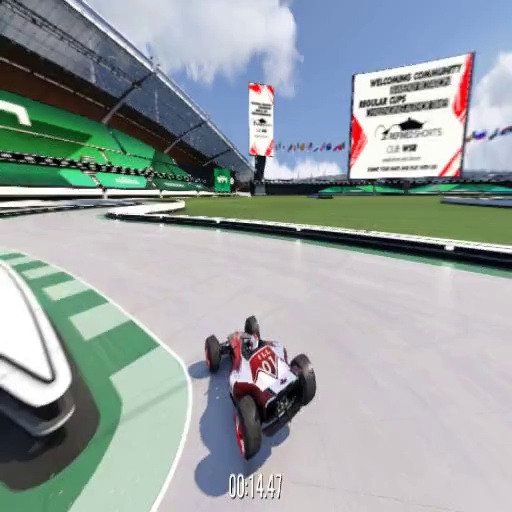}
    \includegraphics[width=0.19\linewidth]{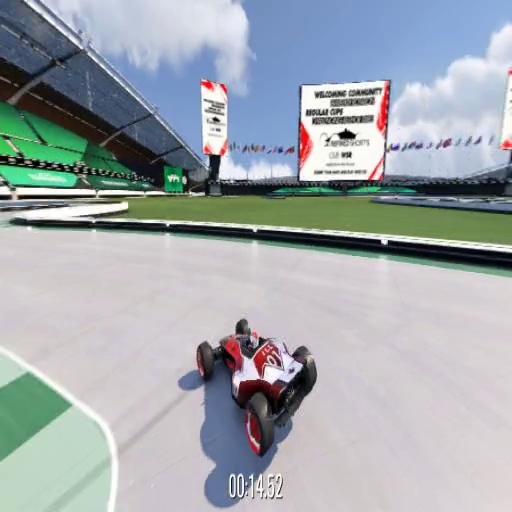}
    \includegraphics[width=0.19\linewidth]{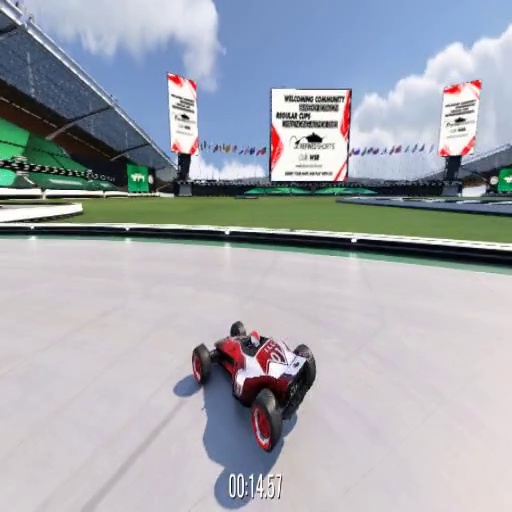}
    \includegraphics[width=0.19\linewidth]{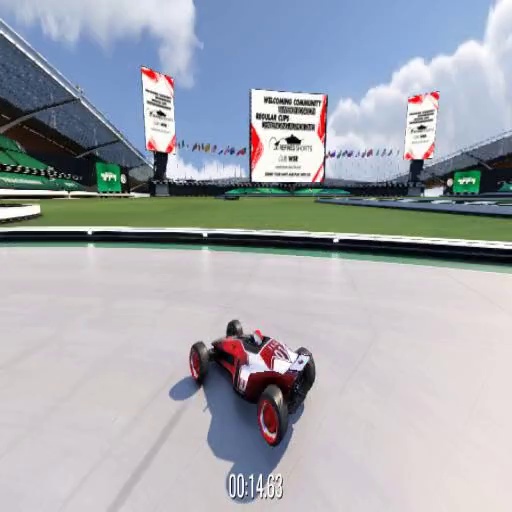}
    \includegraphics[width=0.19\linewidth]{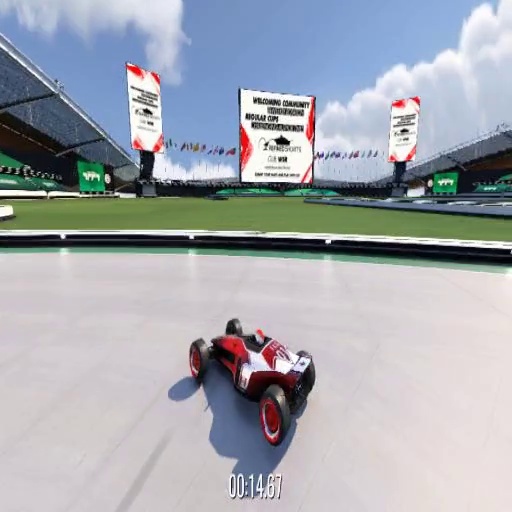}
    \end{center}
    \caption{Some sequences from Trackmania used here for testing.}
    \label{fig:trackmania}
\end{figure}

\textbf{Testing dataset.} For testing purposes we acquired a small dataset of $T = 5500$ frames (approx. 4m 36s), later split into 50\% training and 50\% validation, of Trackmania, a racing video game with clean synthetic visuals (Fig.~\ref{fig:trackmania}).
This dataset was annotated by a human operator answering the three questions in Table \ref{tab:prompt} for each frame $t$.
Manual annotation was reasonably fast: 
while watching a replay played at 10 Hz, the operator was asked to press the space bar when $Q_j$ had a positive answer.
The entire process took less than 1 hour.
The same dataset was used to evaluate plain VLM annotation, to compute the RAM model and for all other experiments reported here.

\textbf{Plain annotation.} Table \ref{tab:base} shows the accuracy, precision, recall, $F1$ and Pearson correlation $\rho$ computed on our dataset for the plain output of different VLMs (Section~\ref{sec:method_data_plain}) at different resolutions, prompted as in Table \ref{tab:prompt}.
We notice a high variability as a function of the metric and question: in our analysis we give therefore the highest importance to $F1$, which is associated with balanced precision and recall. Furthermore, we highlight that a low correlation coefficient with high precision (or recall) identifies degenerate cases where the VLM has a constant output and the dataset is unbalanced.
ChatGPT-5.4 leads on average over all other VLMs, but its Terms of Service forbid downstream AI training. From a practical point of view, thus, Qwen3.5 and Kimi-k2.5 represent better solutions, with Qwen3.5 preferred for $Q_2$ and Kimi-k2.5 providing better answers to $Q_1$ and $Q_3$.
In terms of frame resolution, more pixels lead to better results, but also higher token usage.
For three ChatGTP-5.4, Qwen3.5 and Nemotron2, the same Table also shows the metrics when the model VLMs

\begin{table*}
\caption{VLM annotation comparison. We indicate with (K) VLMs additionally prompted with the per-frame key-press label (W/A/S/D/R). Best results in bold. Tokens per call per question are also reported.}
\label{tab:base}
\scriptsize
\centering
\begin{tabular}{ccc ccccc ccccc ccccc }
\toprule
\multirow{2}{*}{VLM} &
\multirow{2}{*}{Res} &
\multirow{2}{*}{Tkns} &
\multicolumn{5}{c}{Middle ($Q_1$)} &
\multicolumn{5}{c}{Forward ($Q_2$)} &
\multicolumn{5}{c}{Barrier ($Q_3$)} \\
\cmidrule(lr){4-8}\cmidrule(lr){9-13}\cmidrule(lr){14-18}
 & & & Acc & Prec & Rec & F1 & $\rho$ & Acc & Prec & Rec & F1 & $\rho$ & Acc & Prec & Rec & F1 & $\rho$\\
\midrule
\multirow{3}{*}{ChatGPT-5.4} & 512 & 1657 & 80.1 & 83.2 & 79.6 & 81.4 & \textbf{80.0} & 87.7 & 87.8 & 99.6 & 93.3 & 34.2 & 91.4 & 56.8 & \textbf{91.7} & 70.4 & \textbf{74.9}\\
& 256 & 696 & 79.5 & 78.9 & 85.3 & 82.0 & 77.2 & 87.3 & 87.3 & 99.8 & 93.2 & 51.5 & 91.6 & 58.4 & 84.6 & 69.4 & 73.3 \\
& 128 & 456 & 73.9 & 70.7 & 89.2 & 79.0 & 68.8 & 87.2 & 87.3 & 99.7 & 93.1 & 10.3 & 88.4 & 45.5 & 23.6 & 31.2 & 32.9\\
\cmidrule{4-18}
ChatGPT-5.4 (K) & 512 & 1657 & \textbf{87.7} & 86.4 & 90.5 & \textbf{88.5} & 75.3 & 90.9 & \textbf{92.3} & 97.5 & 94.9 & 49.7 & \textbf{95.0} & \textbf{77.7} & 75.7 & \textbf{77.1} & 74.6\\
\cmidrule{2-18}
\multirow{3}{*}{Qwen3.5-397B-A17B} & 512 & 1456 & 75.1 & 72.8 & 86.7 & 79.3 & 62.0 & \textbf{91.5} & 91.2 & 99.7 & \textbf{95.3} & \textbf{59.1} & 87.9 & 44.6 & 34.98 & 39.3 & 38.7 \\
& 256 & 496 & 73.1 & 70.1 & 88.6 & 78.3 & 58.8 & 88.2 & 88.1 & 99.7 & 93.6 & 31.1 & 86.3 & 35.7 & 28.0 & 31.5 & 28.7\\
& 128 & 496 & 74.7 & 71.9 & 87.9 & 79.2 & 60.7 & 87.7 & 87.9 & 99.4 & 93.3 & 27.3 & 87.4 & 41.2 & 30.5 & 35.2 & 34.2\\
\cmidrule{4-18}
Qwen3.5-397B-A17B (K) & 512 & 1456 & 79.4 & 76.1 & 90.5 & 82.8 & 58.8 & 89.0 & 89.7 & 98.6 & 94.0 & 36.7 & 90.4 & 63.1 & 32.5 & 43.2 & 41.4\\
\cmidrule{2-18}
\multirow{3}{*}{Nemotron3-Omni} & 512 & 1665 & 74.7 & 68.5 & \textbf{99.4} & 81.1 & 57.9 & 87.1 & 87.2 & 99.6 & 93.0 & 4.8 & 88.7 & 10.7 & 0.2 & 0.4 & 1.0\\
& 256 & 1665 & 77.5 & 71.1 & 98.9 & 82.8 & 62.0 & 86.9 & 87.2 & 99.4 & 92.9 & 3.3 & 88.7 & 10.7 & 0.2 & 0.5 & 1.0\\
& 128 & 1665 & 77.3 & 71.2 & 98.0 & 82.6 & 61.3 & 87.0 & 87.1 & 99.7 & 93.0 & $-$1.8 & 89.4 & 65.9 & 9.1 & 16.1 & 26.8 \\
\cmidrule{2-18}
\multirow{3}{*}{Nemotron2-12B} & 512 & 434 & 66.1 & 62.0 & 97.9 & 76.0 & 39.9 & 87.2 & 87.1 & \textbf{99.9} & 93.1 & --- & 89.7 & 70.3 & 12.4 & 21.3 & 31.1\\
& 256 & 434 & 65.8 & 61.8 & 98.3 & 75.6 & 39.9 & 87.2 & 87.1 & \textbf{99.9} & 93.1 & --- & 89.6 & 61.3 & 18.0 & 28.0 & 33.9\\
& 128 & 434 & 67.7 & 63.4 & 96.6 & 76.7 & 42.1 & 87.2 & 87.1 & \textbf{99.9} & 93.1 & --- & 90.6 & 65.5 & 31.9 & 43.1 & 47.1\\
\cmidrule{4-18}
Nemotron2-12B-VL (K) & 512 & 1665 & 61.0 & 58.6 & 97.9 & 73.4 & 25.2 & 88.4 & 89.4 & 98.2 & 93.6 & 32.3 & 89.8 & 64.5 & 17.2 & 27.4 & 30.5\\
\cmidrule{2-18}
Kimi-k2.5 & 512 & 1979 & 79.8 & \textbf{92.6} & 76.7 & 84.0 & 65.7 & 68.0 & 73.1 & 80.9 & 76.9 & 30.3 & 94.6 & 75.0 & 43.2 & 55.3 & 60.5\\
\bottomrule
\end{tabular}
\end{table*}

\begin{table*}
\caption{RAM annotation with different VLMs and a mix (ChatGPT-5.4 + Qwen3.5-397B-A17B + and Nemotron3-Omni) of them. Snorkel~\cite{ratner2017snorkel} results on the same mix are also reported.}
\label{tab:ram}
\scriptsize
\centering
\begin{tabular}{ccc ccccc ccccc ccccc }
\toprule
\multirow{2}{*}{VLM} &
\multirow{2}{*}{Res} &
\multirow{2}{*}{Tkns} &
\multicolumn{5}{c}{Middle ($Q_1$)} &
\multicolumn{5}{c}{Forward ($Q_2$)} &
\multicolumn{5}{c}{Barrier ($Q_3$)} \\
\cmidrule(lr){4-8}\cmidrule(lr){9-13}\cmidrule(lr){14-18}
 & & & Acc & Prec & Rec & F1 & $\rho$ & Acc & Prec & Rec & F1 & $\rho$ & Acc & Prec & Rec & F1 & $\rho$\\
\midrule
ChatGPT-5.4 & 512 & 1657 & 88.3 & 84.6 & 95.8 & 89.9 & 77.3 & 87.2 & 87.1 & \textbf{99.9} & 93.1 & 1.2 & 94.5 & 72.2 &	80.0 & \textbf{76.2} & \textbf{75.4}\\
ChatGPT-5.4 (K) & 512 & 1657 & 87.7 & 85.6 & 91.8 & 88.7 & 75.6 & 90.9 & \textbf{92.5} & 97.3 & 94.9 & 52.2 & 94.5 & 73.0 & 77.4 & 75.5 & 74.3\\
Qwen3.5-397B-A17B & 512 & 1456 & 77.9 &	73.9 & 91.9 & 82.0 & 56.8 & 
\textbf{91.7} & 91.4 & 99.6 & \textbf{95.4} & \textbf{57.5} & 90.8 & 61.4 & 44.6 &	51.9 & 48.3\\
Qwen3.5-397B-A17B (K) & 512 & 1456 & 78.0 & 72.9 & 94.9 & 82.5 & 58.2 & 88.3 & 88.6 & 99.1 & 93.6 & 29.1 & 90.4 & 58.0 & 49.4 & 53.6 & 49.1\\
Nemotron3-Omni & 512 & 1665 & 79.2 & 72.8 & 98.7 & 83.9 &	61.7 & 87.1 & 87.1 & 99.8 & 93.1 & -1.2 & 91.1 & 56.9 & 81.9 & 67.4 & 64.1 \\
Nemotron3-Omni (K) & 512 & 1665 & 59.3 & 57.4 & \textbf{99.8} & 72.9 & 23.7 & 88.7 & 89.2 & 98.7 & 93.8 & 34.0 & 88.9 & 50.3 & 25.2 & 33.7 & 32.3\\
Nemotron2-12B & 512 & 434 & 68.7 & 64.1 & 97.3 & 77.3 & 41.7 & 89.3 & 91.9 & 96.0 & 94.0 & 46.4 & 90.7 & 59.3 & 51.9 & 55.6 & 50.9\\
Kimi-k2.5 & 512 & 1979 & 86.9 & \textbf{91.3} & 89.3 & \textbf{90.4} & 70.0 & 65.8 & 65.8 & 99.8 & 79.4 & --- & \textbf{95.0} & \textbf{81.0} & 44.2 & 57.6 & 58.7\\
\midrule
Mix & 512 & 4778 & \textbf{89.0} & 86.0 & 95.1 & \textbf{90.4} & 78.2 &
91.3 & 91.4 & 99.1 & 95.2 & 56.0 & 94.2 & 74.3 & 71.9 & 73.4 & 71.3\\
\midrule
Snorkel-MV & 512 & --- & 65.4 & 61.7 & 97.2 & 75.5 & 38.6 & 75.4 & 86.2 & 85.5 & 85.8 & $-$8.9 & 23.3 & 12.2 & \textbf{93.9} & 21.5 & 9.5\\
Snorkel-LM & 512 & --- & 88.4 & 84.2 & 97.1 & 90.2 & \textbf{78.9} & 64.2 & 90.6 & 65.8 & 76.2 & 12.7 & 25.5 & 0.0 & 0.0 & 0.0 & $-$47.2\\
\bottomrule
\end{tabular}
\end{table*}

\textbf{RAM evaluation.} Table \ref{tab:ram} shows significant improvements provided by RAM over plain annotation for any VLM considered.
Furthermore, RAM reduces the gap between different VLMs.
ChatGPT-5.4 remains the best model on average, but with RAM, Qwen3.5 and Nemotron2-12B can also answer all three questions with reasonable $F1$ and $\rho$.
Table ~\ref{tab:ram} also compares RAM against two Snorkel-style aggregators
on ChatGPT-5.4 output.
Snorkel-LM ties RAM on $Q_1$ ($F1 = 90.2$, $89.9$ for RAM) but collapses on the sparse $Q_3$ ($F1 = 0.0$, $76.2$ for RAM), confirming that RAM's supervised aggregation strategy is superior when classes are imbalanced; Snorkel-MV underperforms across all questions.

\begin{figure*}[t]
  \centering  \includegraphics[height=0.44\columnwidth,trim={0.2cm 0.2cm 3.8cm 0.2cm}, clip]{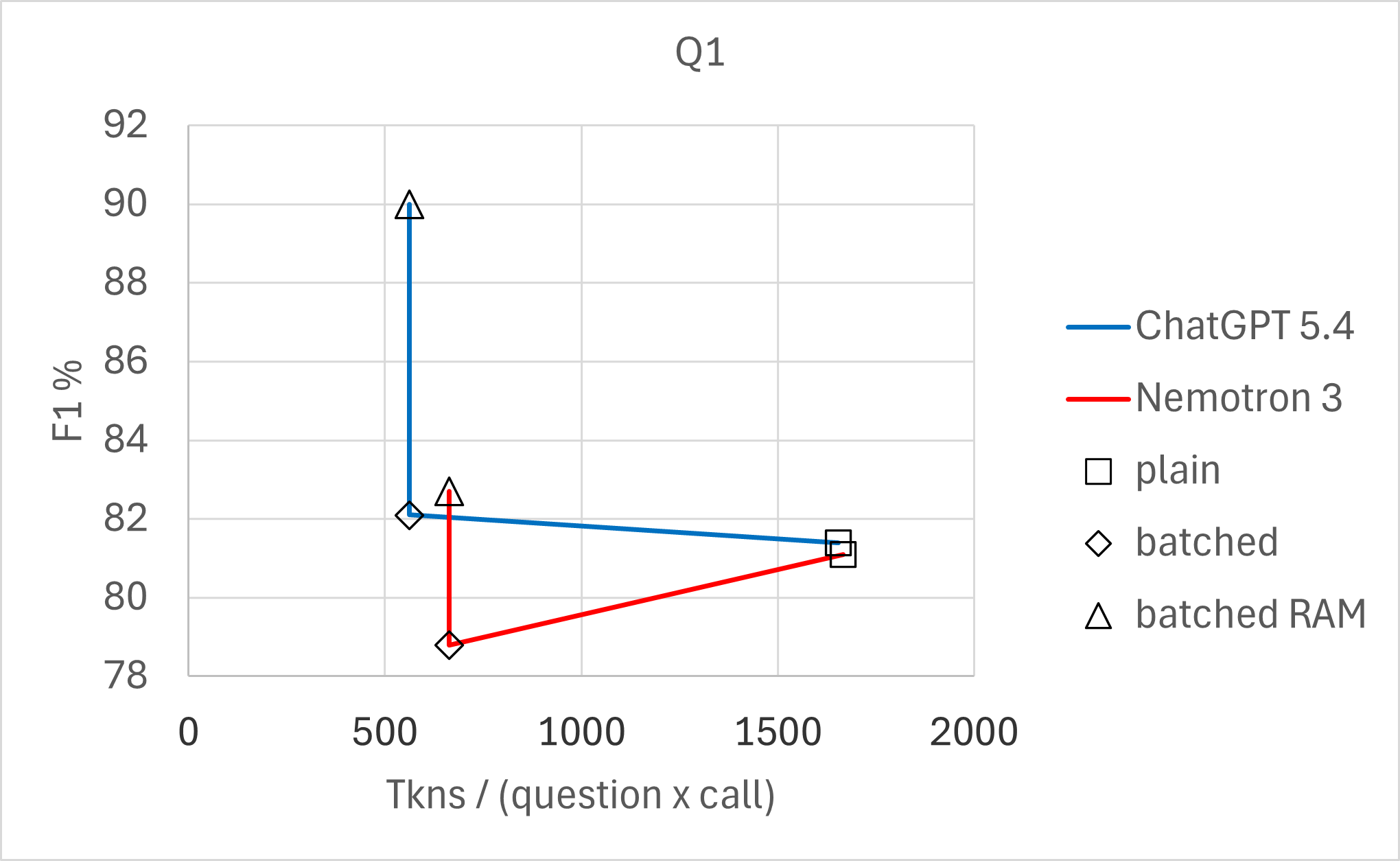}  \includegraphics[height=0.44\columnwidth,trim={0.2cm 0.2cm 3.8cm 0.2cm}, clip]{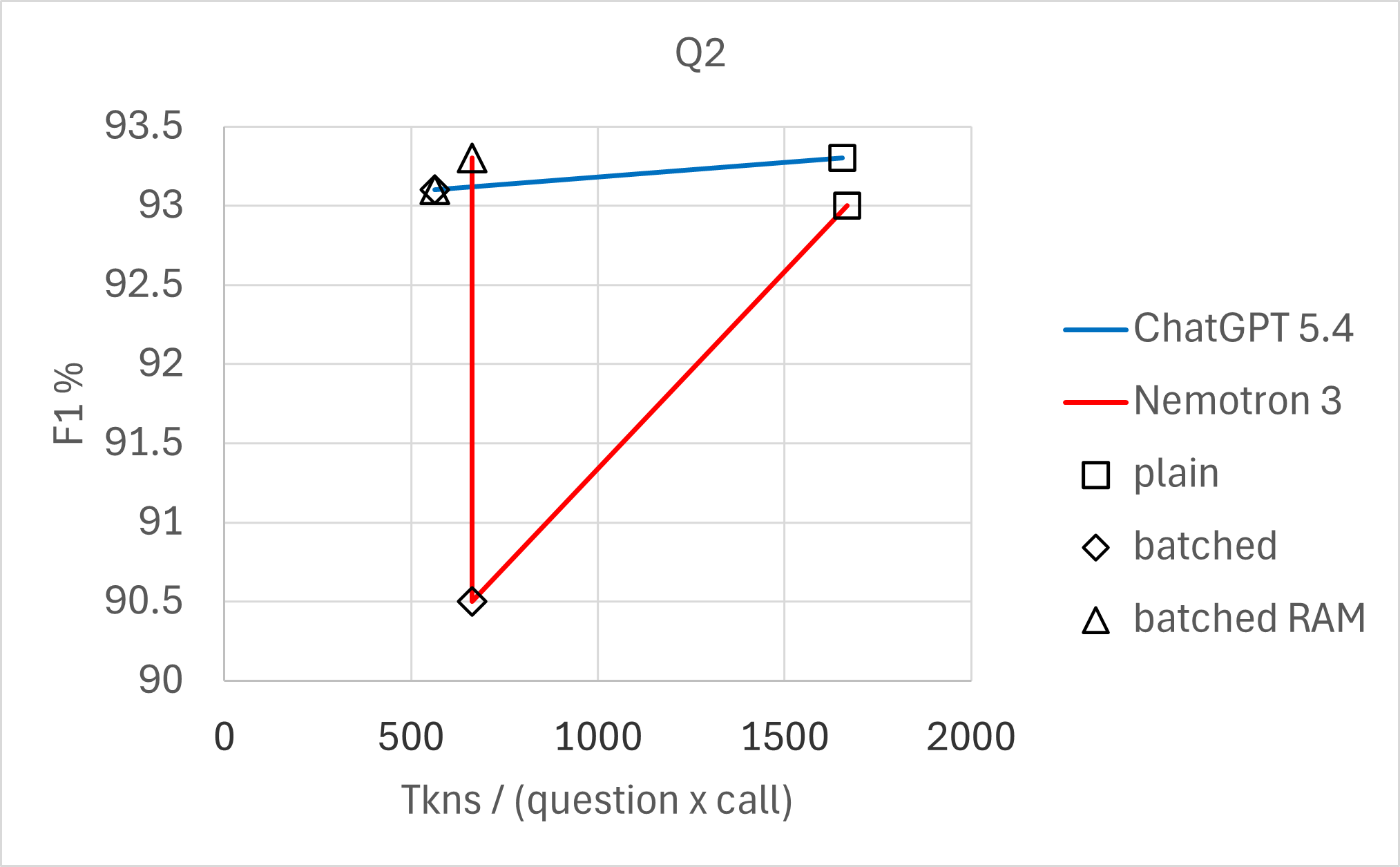}  \includegraphics[height=0.44\columnwidth,trim={0.2cm 0.2cm 0.2cm 0.2cm}, clip]{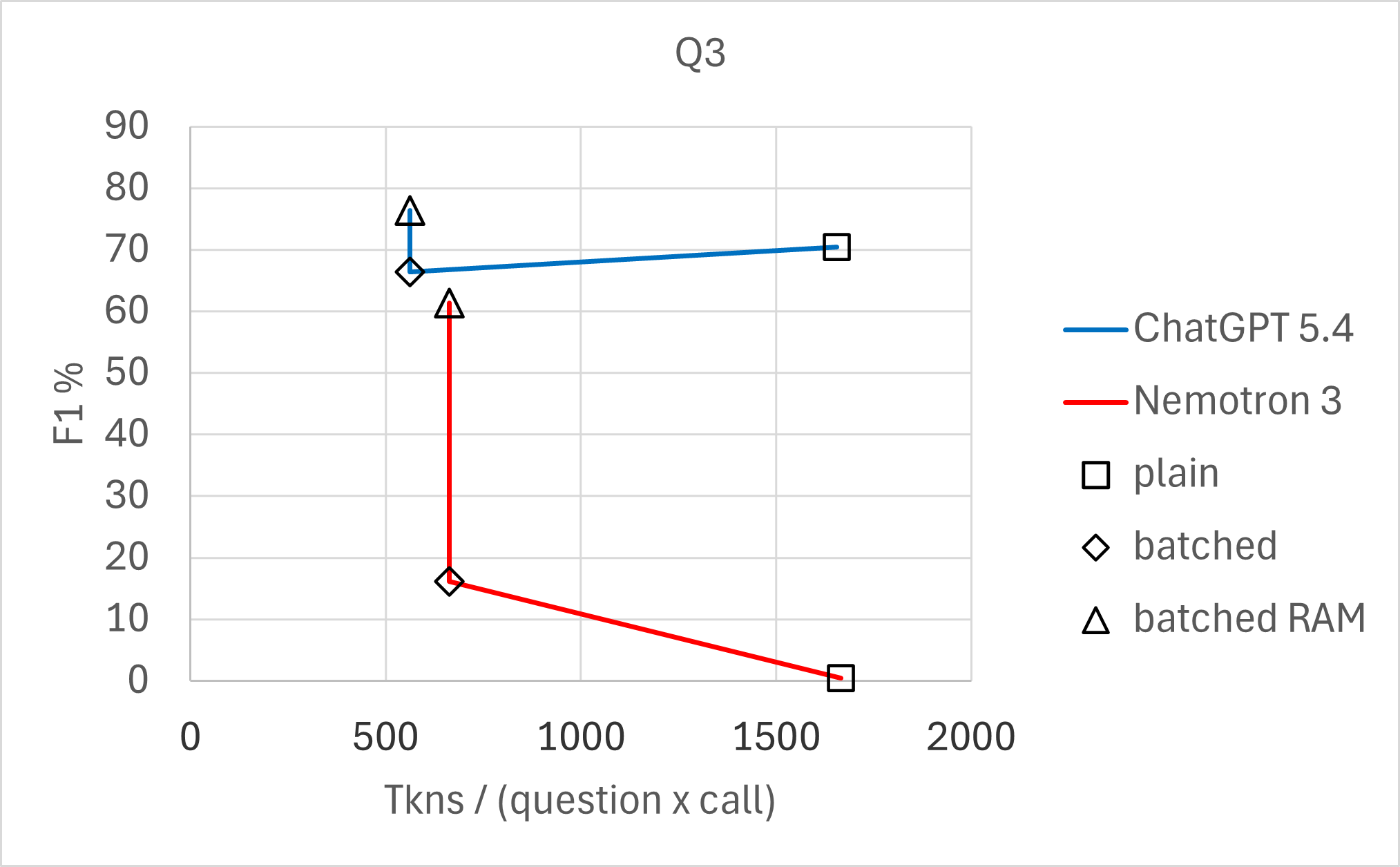}
    \caption{F1 vs. the number of tokens per question per call for ChatGPT and Nemotron3-Omni and the questions in Table \ref{tab:prompt} in plain configuration or using batching (Section \ref{sec:batching}) and batching + RAM (Section \ref{sec:method_data_plain}).}
  \label{fig:ablation_1}
\end{figure*}

\begin{figure*}[t]
  \centering  \includegraphics[height=0.44\columnwidth,trim={0.2cm 0.2cm 3.8cm 0.2cm}, clip]{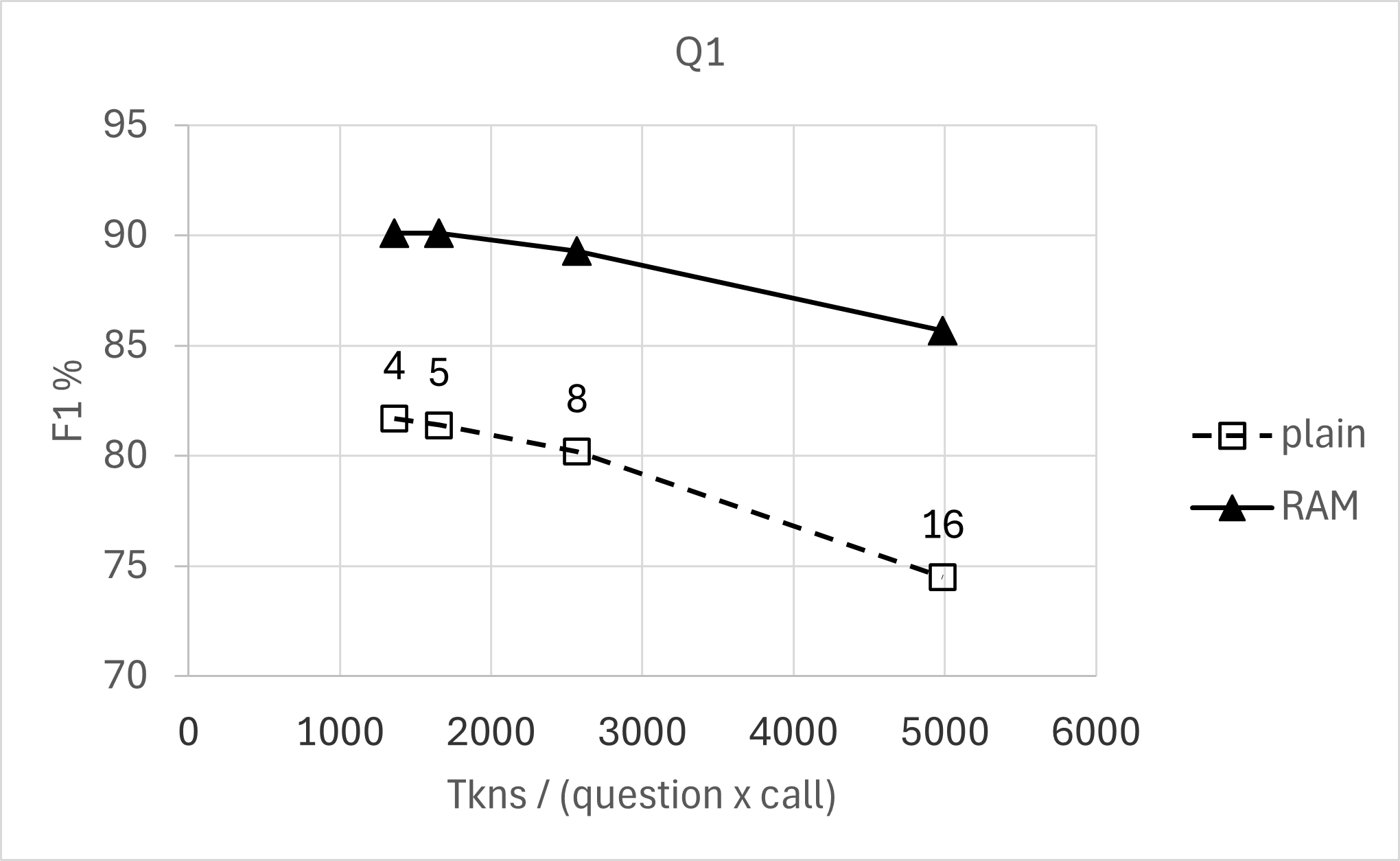}  \includegraphics[height=0.44\columnwidth,trim={0.2cm 0.2cm 3.8cm 0.2cm}, clip]{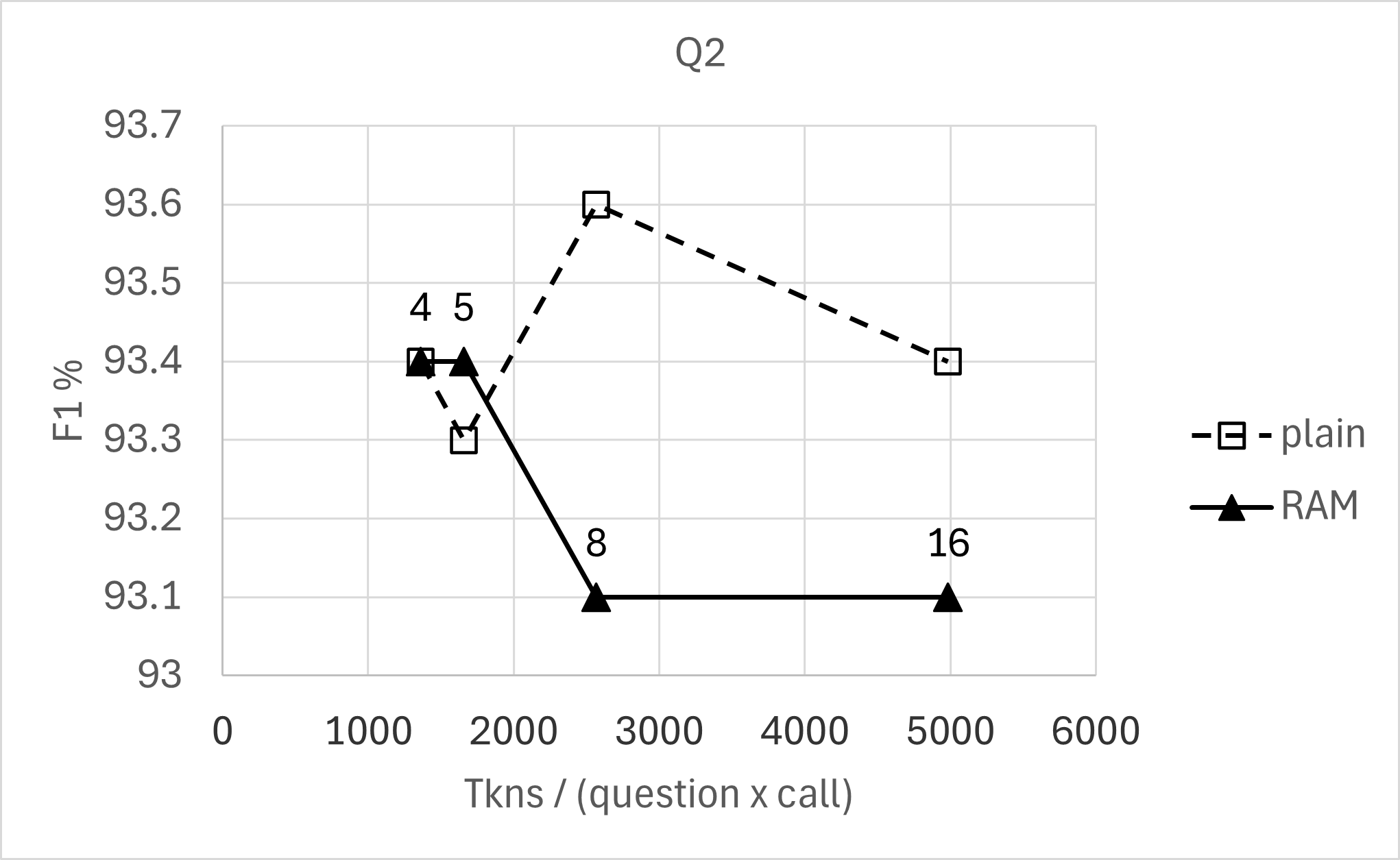}  \includegraphics[height=0.44\columnwidth,trim={0.2cm 0.2cm 0.2cm 0.2cm}, clip]{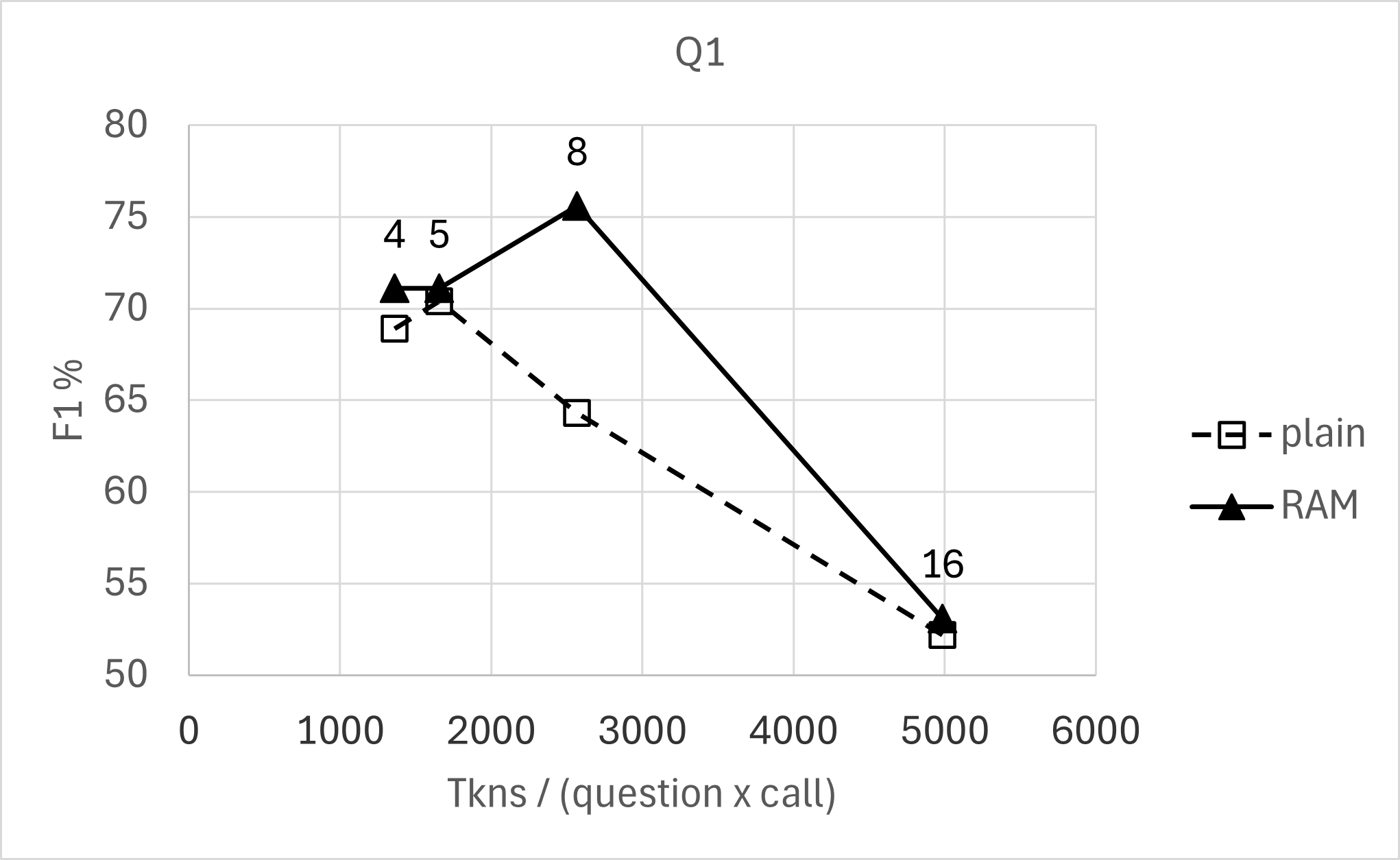}
  \caption{F1 vs. the number of tokens per question per call for ChatGPT and Nemotron3-Omni and the questions in Table \ref{tab:prompt} in plain configuration and RAM for ChatGPT-5.4 and various frame sequence lengths in input to the model (indicated as labels in the graphs).}
  \label{fig:ablation_2}
\end{figure*}

\textbf{Question batching}\label{sec:batching}
 Fig.~\ref{fig:ablation_1} shows the effect of question batching on $F1$ and the number of required tokens.
Batching reduces the tokens, but interference between the questions in the same prompt affects the quality of annotation, generally reducing $F1$.
Interestingly enough, applying RAM to the batched questions pushes $F1$ up again, achieving the best compromise between quality and token usage.
Fig.\ref{fig:ablation_2} shows $F1$ as a function of the length of the input sequence: longer input sequences require more tokens per annotation with $F1$ decreasing, likely because the correct answer to the given question may vary per frame along the sequence itself, thus decreasing the confidence of the VLM.

\textbf{Prompt optimization.}
Our experiments with MIPROv2 delivered contrasting results: Table~\ref{tab:nothink_results} shows that $F1$ improves for $Q_1$ and $Q_3$, but the original prompt is better for $Q_1$ after RAM. The effect of MIPROv2 on $Q_2$ is negligible, but it leads to some improvement after RAM.
Overall we found the output of MIPROv2 to be poorly predictable especially in cases of unbalanced datasets, as for $Q_2$ and $Q_3$.
Addressing these failure modes — e.g. through class-balanced few-shot sampling, per-question reweighting of the optimization metric, or hard-example mining — is left as future work.


\textbf{Wreckfest.} Wreckfest (Fig.~\ref{fig:teaser}) is a demolition-derby game with complex multi-car scenes and realistic damage rendering. Table \ref{tab:wreckfest} shows F1 for the plain and RAM output of different VLMs on Wreckfest sequences.
It confirms that RAM consistently improves the annotation quality, but comparison against Tables~\ref{tab:base} and \ref{tab:ram} highlights that VLMs may struggle on video games with richer visual features (when compared to Trackmania) or less favorable camera configuration (first-person view as for Wreckfest in Fig.~\ref{fig:teaser} vs. third-person view as for Trackmania in Fig.~\ref{fig:trackmania}). 

{\color{red}
}

\begin{table}[t]
\caption{$F1$ and $\rho$ with and without prompt optimization.}
\label{tab:nothink_results}
\scriptsize
\centering
\begin{tabular}{l l c c c c}
\toprule
Question & Variant  & Plain F1 & RAM F1 & $\rho$ \\
\midrule
\multirow{2}{*}{Middle}
  & ChatGPT-5.4 (base)   & 78.7          & \textbf{90.3} & \textbf{77.8} \\
  & ChatGPT-5.4 (MIPROv2)   & \textbf{86.4} & 86.5          & 71.0 \\
\midrule
\multirow{2}{*}{Forward}
  & ChatGPT-5.4 (base)    & \textbf{93.2} & 93.1          & ---  \\
  & ChatGPT-5.4 (MIPROv2)   & 93.1          & \textbf{94.9} & \textbf{50.1} \\
\midrule
\multirow{2}{*}{Barriers}
  & ChatGPT-5.4 (base)    & 57.0          & 69.8          & 66.3 \\
  & ChatGPT-5.4 (MIPROv2) &  \textbf{59.4} & \textbf{73.9} & \textbf{71.1} \\
\bottomrule
\end{tabular}
\end{table}


\begin{table*}
\caption{F1 for plain and RAM annotation on WRECKFEST.}
\label{tab:wreckfest}
\scriptsize
\centering
\begin{tabular}{cc cc cc cc}
\toprule
\multirow{2}{*}{VLM} &
\multirow{2}{*}{Tkns} &
\multicolumn{2}{c}{Middle ($Q_1$)} &
\multicolumn{2}{c}{Forward ($Q_2$)} &
\multicolumn{2}{c}{Barrier ($Q_3$)} \\
\cmidrule(lr){3-4}\cmidrule(lr){5-6}\cmidrule(lr){7-8}
 & & plain & RAM & plain & RAM & plain & RAM\\
\midrule
ChatGPT-5.4 & 669 & 37.5 & 55.4 & 89.4 & 83.1 & 15.7 & 21.9 \\
Nemotron3-Omni & 1949 & 9.7 & 50.7 & 80.4 & 81.5 & 13.3 & 14.3 \\
Qwen3.5-397B-A17B & 467 & 41.9 & 46.4 & 74.1 & 75.1 & 11.4 & 12.3 \\
\bottomrule
\end{tabular}
\end{table*}
\section{Discussion and Conclusion}

We studied reward annotation for video games through VLMs and shown that the VLM choice, question batching, frame resolution and sequence length impact the annotation quality and the token consumption.
By leveraging the correlation between the VLM answers (via a simple linear model like RAM) we find a better compromise between the annotation cost and its quality.
Other approaches, like automatic prompt optimization, provide less consistent improvements.

As for the limitations, the significance of our tests is so far limited by the small single-annotator reference set: this is why we plan to  study larger sets and more games.
Another important factor comes from the commercial constraints of some VLMs whose terms of service prevent downstream AI training; furthermore, content filters can block violent scenes ---  broadly defined from car accidents to war scenes ---  that are common in gaming.
The fact that VLMs still struggle on complex video game scenarios, as shown for Wreckfest, suggests that significant room for improvement still exists and more research is needed to use VLMs as reliable video game annotators.

\bibliographystyle{plain} 
\bibliography{references} 

\end{document}